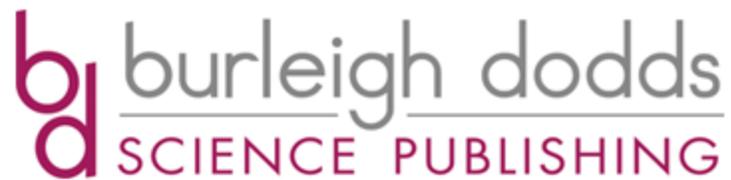

**CH10 - Advances in soft robotics in agriculture**

| Publisher: | *Burleigh Dodds Science Publishing* |
|---|---|
| Manuscript ID | BDSP-BK-2022-0167.R2 |
| Manuscript Type: | Book |
| Date Submitted by the Author: | 20-Mar-2023 |
| Complete List of Authors: | Leylavi Shoushtari, Ali; Wageningen University & Research, Farm Technology Group, Plant Sciences Department |
| Keywords: | Soft Grippers, Agricultural robotics and automation, Soft actuators, Shape adaptability, Agro-food robotics |
| | |

SCHOLARONE™
Manuscripts

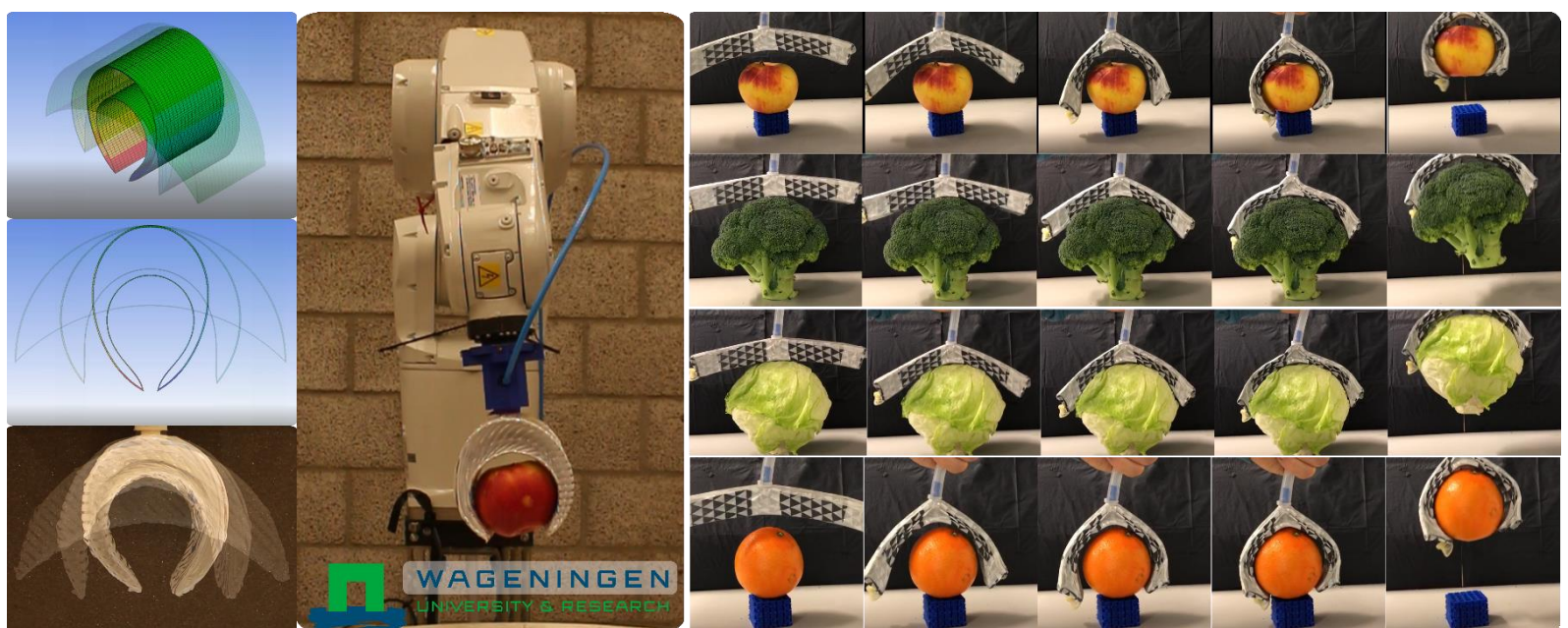

# Advances in soft grasping in agriculture

**Ali Leylavi Shoushtari**

1-Wageningen Robotics, Wageningen, The Netherlands

ali.leylavishoushtari@wagrobotics.com

2-Farm Technology Group, Wageningen University & Research, Wageningen, The Netherlands

**Abstract**

Agricultural robotics and automation are facing some challenges rooted in the high variability of products, task complexity, crop quality requirement, and dense vegetation. Such a set of challenges demands a more versatile and safe robotic system. Soft robotics is a young yet promising field of research aimed to enhance these aspects of current rigid robots which makes it a good candidate solution for that challenge. In general, it aimed to provide robots and machines with adaptive locomotion (Ansari et al., 2015), safe and adaptive manipulation (Arleo et al., 2020) and versatile grasping (Langowski et al., 2020). But in agriculture, soft robots have been mainly used in harvesting tasks and more specifically in grasping. In this chapter, we review a candidate group of soft grippers that were used for handling and harvesting crops regarding agricultural challenges i.e. safety in handling and adaptability to the high variation of crops. The review is aimed to show why and to what extent soft grippers have been successful in handling agricultural tasks. The analysis carried out on the results provides future directions for the systematic design of soft robots in agricultural tasks.

## 1. Introduction

Despite remarkable progress in agri-food robotics, there still no commercially available autonomous harvester due to the scientific and technological gaps (Adamides & Edan, 2023; Bac et al., 2014; Droukas et al., 2023; Kumar & Mohan, 2022; Martin et al., 2022; Wang et al., 2022). In the process of automating agricultural tasks, robotic technologies are facing a wide variety of challenges. From a robotics point of view, these challenges could be classified as perception and action. The core of the perception challenge is to comprehend, reason, and make sense of the unknown and cluttered environment surrounding the robot e.g. arable farms or greenhouses. Whereas the challenges of action (CoA) are about agile maneuvering in and physically interacting with an unstructured environment characterized by high diversity and uncertainty. From a robotics point of view, the former class of challenges is being formulated as computer vision and artificial intelligence (AI) problems while the latter is framed as manipulation, grasping, and navigation problems. Six decades of research on computer vision and AI (Szeliski, 2011) made it mature enough to be applied to agricultural tasks. Meanwhile, robotic grasping, manipulation, and navigation still require further research.

Conventional robots are usually made of rigid materials that limit their capability in adapting to objects and obstacles. Furthermore, physical interaction of robots with the environment is inevitable whether in locomotion and manipulation or even in form of collision and the mentioned limited adaptability makes rigid robots even unsafe. In particular, these features i.e. adaptability and safety become the limiting factors for rigid robots, especially when they are implemented in a real-world scenario and are interacting with an unstructured environment with high diversity and uncertainty as is the case in agriculture. Furthermore, maintaining the quality of products that the robotic systems are handling is a significant requirement in agricultural tasks e.g. harvesting with rigid robots (in this case grippers) seemed to not be a proper option for automation.

Soft robots could be defined as robots and machines that are "primarily composed of easily deformable matter such as fluids, gels, and elastomers that match the elastic and rheological properties of biological tissue and organs" (Majidi, 2013). Although there are machines and robots made of materials that have remarkably less elasticity compared to elastomers but still can provide large deformations due to their particular geometry. Compliant mechanisms are an example of this group of robots or mechanisms (Howell et al., n.d.). The change of the material from rigid to deformable and stretchable provides robots and machines with the adaptability of shape and a locomotion or manipulation strategy for a broad range of tasks and environments (Majidi, 2013). Soft Robotics is a relatively young yet promising field of research aimed to enhance robots' safety and adaptability while keeping them simple and affordable. In agricultural robotics, crops are usually vulnerable and delicate, have high variation (i.e. different shape, size, surface texture and roughness, and softness), and usually are affected by environmental conditions e.g. moisture and dust which makes them challenging to handle using conventional rigid robots. The above-mentioned features of soft robotics can significantly contribute to the robots' versatility, making it an ideal solution for agricultural robotic challenges. To elucidate this, this chapter aims to answer the question of how and to what extent soft robotics can enhance the versatility of robots, especially in grasping and manipulation.

Section 1 will focus on the type of challenges that any robotic system would face in agricultural tasks. Section 2 which is the main part of this chapter is dedicated to how soft robotics could help us to deal with these challenges. In section 3 a case study will be presented which in a nutshell reflects how soft agricultural robotic research is carried out. Section 4 is a summary of this chapter and section 5 is a short list of valuable information about agricultural robotics including review papers and open-source websites.

**1.1. The Challenges of Action (CoA) in agricultural robots are:**
- **High diversity in agricultural products:** Pose (position and orientation), shape, size, mechanical properties (structural stiffness), surface properties i.e. roughness and adhesion of crops.

    **Diversity between different crop variants:** Even crops that genetically belong to the same family can have different variants with quite diverse shapes and sizes. For instance, cucurbit crops can have several variants e.g. cucumber, melon, watermelon, and squash which have completely different shapes and sizes (Figure 1 A-D).

    **Diversity within a crop variant (inherent diversity):** Although the crops within a variant look similar but still are not identical. This diversity is also time-dependent and dynamic. Figure 2-A shows the shape and size diversity of 11 different variants of cucurbits. The quantified dynamics of size variation and shape variation could be seen in Figure 2B-C and figure 2-D respectively.

- **Safety requirements in the handling of crops:** in an automated farm, crops are commercial products, hence their quality should be guaranteed by the machines that are interacting with them. It means, that either during crop maintenance or harvesting, robots are required to satisfy a high level of safety.
- **Task complexity:** agricultural tasks are multifaceted. For instance the harvesting task requires not only grasping objects but also detaching, holding them robustly while confronted with dynamic perturbances during manipulation. This challenge demands the robots i.e. grippers to be able to deal with a high variation in force and torques.
- **Dense vegetation environments:** such an environment challenges the maneuvering of robotic systems. There, a robot with a bulky body cannot be agile and maneuverable enough to finish its task within an economically feasible amount of time while avoiding collisions i.e. self-body collisions and collisions with plants.

Advances in agri-food robotics (ed. Prof Eldert van Henten and Prof Yael Edan)

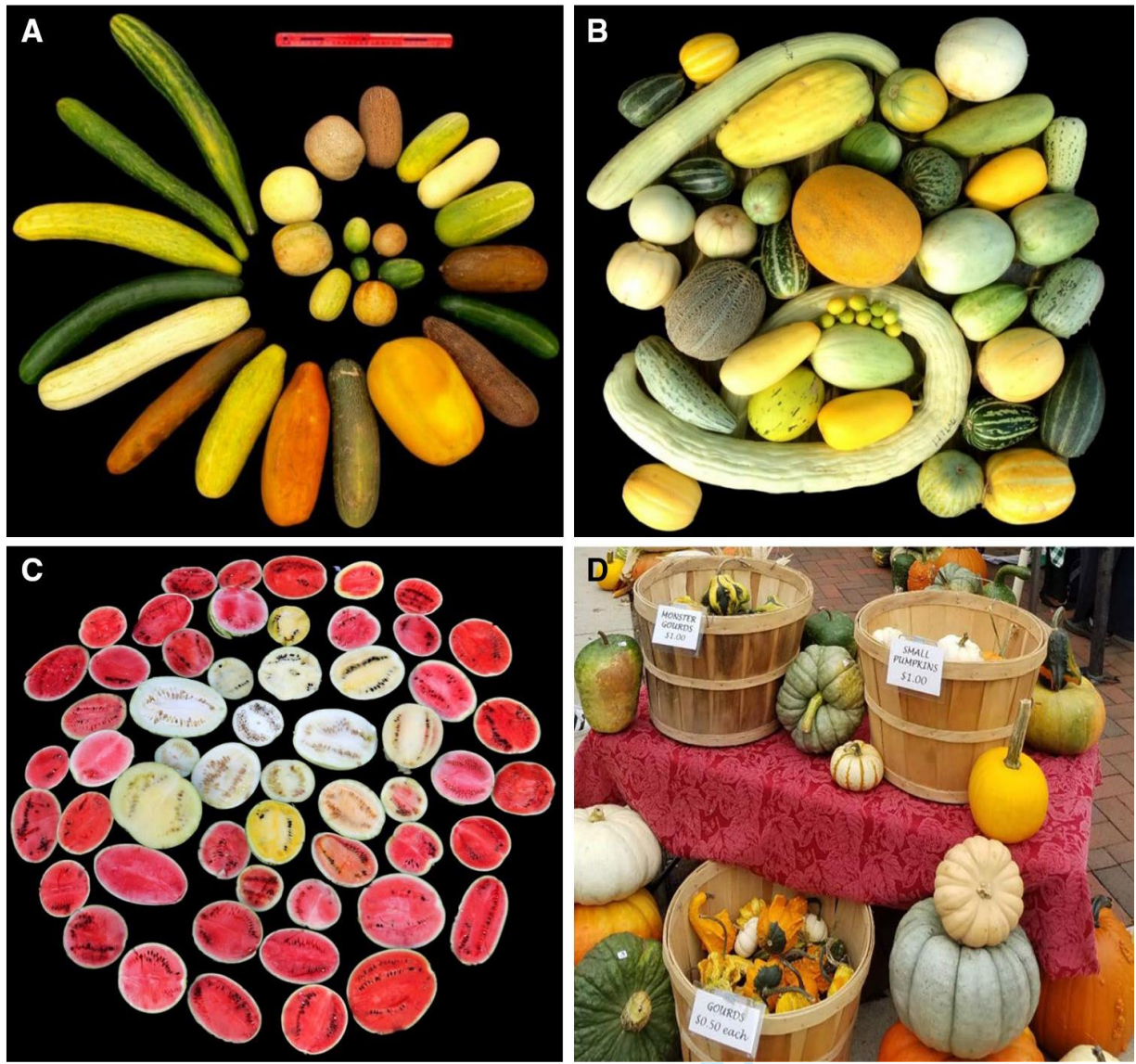

Figure 1 High diversity of shape and size of cucurbit crops. A-D depicts the diversity within different cucurbit variants i.e. cucumber, melon, watermelon, and squash. This figure is used directly from (Pan et al., 2020).

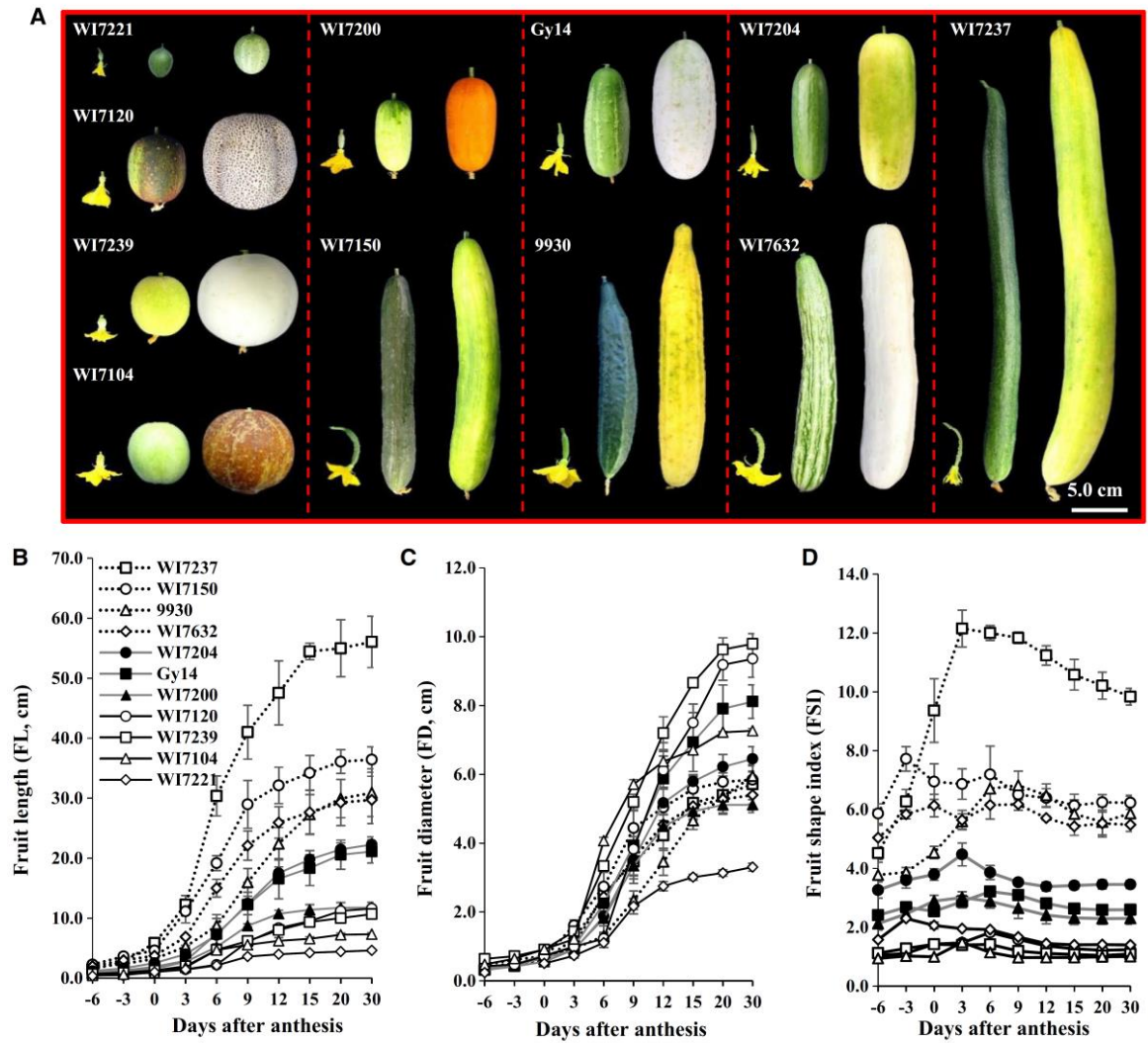

Figure 2 Growing dynamics of cucumbers with different genetic backgrounds. A - images of ovary and fruit of 11 cucumbers depicting the dynamic of growth at 0, 12, and 30 days after the pollination. B-C and represent the quantifications of size and shape dynamics respectively. This figure is used directly from (Pan et al., 2020)

## 2.  Soft Agricultural Robots: how soft robotics can address the agricultural challenges.

Soft robotics is a relatively new field of research aimed to make robots safer and more adaptive. High adaptability is useful, particularly for robotic grasping where robots are required to conform to objects with high variations (challenge 1). Moreover, when the objects to be handled are delicate/fragile then safety (in handling) would also be an important requirement (challenge 2). In summary both safety and adaptability are the main requirements of agri-food robotics that increase the demand for soft grippers.

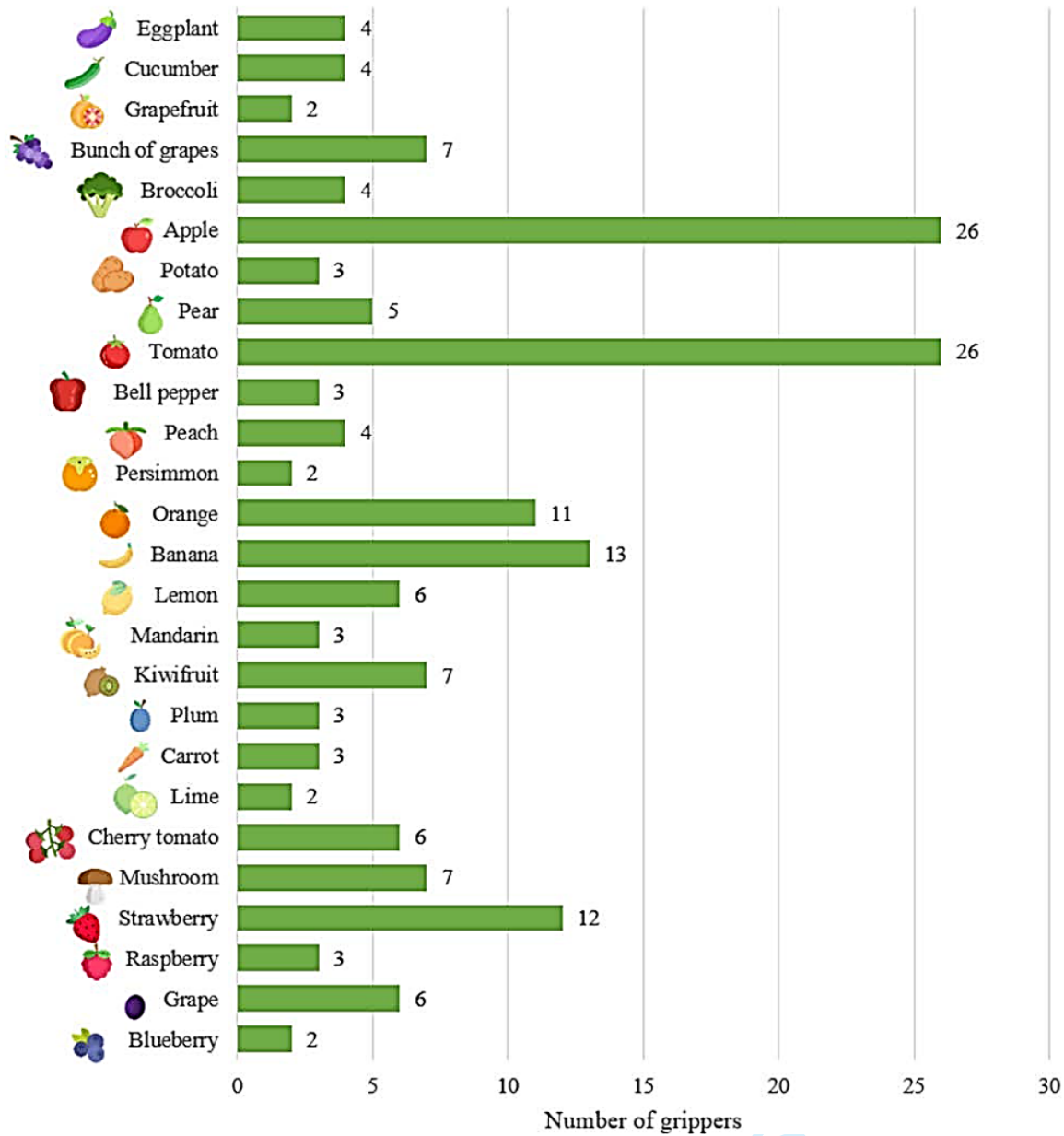

Figure 3 Number of soft grippers used to harvest various crops. Adopted from *(Elfferich, 2022)*

During the last three decades a large number of soft robotic grippers have been designed and developed for crop harvesting and handling (Figure 3) mainly focused on the 2 above-mentioned motivations. Meanwhile, there are two further challenges (3 and 4) for a soft gripper to meet to be suited for agricultural tasks e.g. greenhouse harvesting. Despite their high adaptability, most soft grippers are not strong enough to handle objects of various weights and are not able to deal dynamically with perturbations in load during harvesting (challenge 3). Although there are some soft grippers with enough strength to deal with the variable loads, this strength is usually high along on axis of the gripper while it is low along other axes which make the gripper flexible along those axes. For instance, considering figure 4-B as a schematic representation of the strength of the soft grippers, usually the gripper has high strength along Y axis and is adaptive under lateral forces applied along the X and Z axis.

In particular, a versatile robotic gripper in agriculture is required to be able to exert enough torques and forces on crops in order to hold, detach and manipulate them. Figure 4-A shows such requirements for a manual harvesting process. Rigid grippers in slippage-free cases are more successful to apply such forces and torques since their rigid links or bodies can endure the forces and wrenches received during physical interaction with objects. Whereas most soft grippers tend to stiffen in a just one direction or plane (i.e. the plane where bending happens and lack the supporting stiffness in other directions or planes). As a result, their strength is maximized only in the bending direction or plane and minimized in the other planes. Soft grippers, therefore, bend once they undergo lateral loads and wrenches induced by the object weight or during pulling, flicking, twisting, bending or accelerating, and deaccelerating. Figure 4-B provides a schematic representation of the heterogenous strength of soft grippers where the strength is maximized just along the grasping direction i.e., Y in this example.

## 2.1 An assessment matrix for soft agri-food grippers

The abovementioned issue with soft grippers has been addressed in a recent work (Figure 4-C) where researchers added an adaptive rigid bar mechanism to the soft fingers to limit the lateral bending of the finger and as a result obtained higher strength in the bending direction also (Zhu et al., 2022). Although these types of solutions have remarkably increased the strength of soft grippers, it has also made them larger and heavier due to the size and weights of the components (i.e. mechanical mechanisms and their actuators) added to the soft gripper. In contrast, challenge number 4 demands thin and lightweight robotic grippers. Although there exist ultra-thin, high-strength, and even ultra-fast soft grippers (Shintake et al., 2016a) they are challenged by the issue of strength heterogeneity (Figure 4-D).

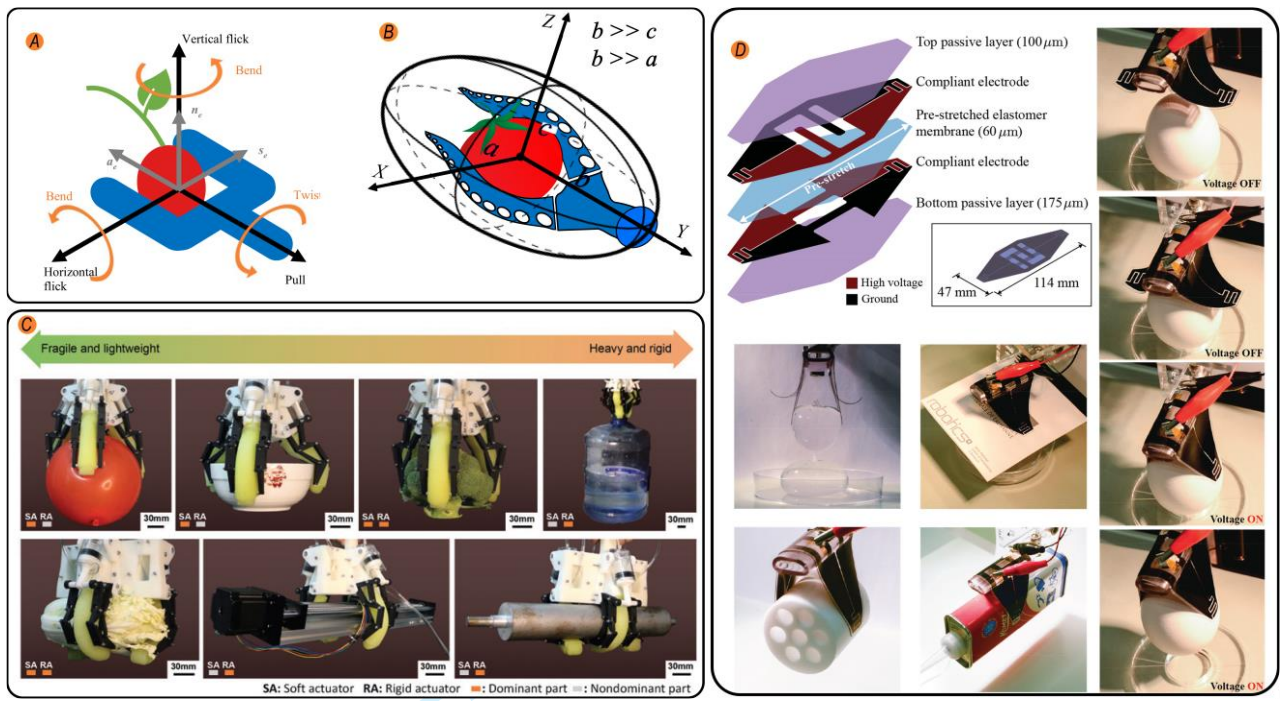

Figure 4 A- Different ways of detaching crops from peduncle (adopted from *(Elfferich, 2022)* ) B- hypothetical ellipsoid of the strength of a soft gripper where a, b and c are representing the strength of the gripper in 3 different directions. The maximum strength is always in the direction of the bending (Y). D- is an example of a soft gripper that can apply grasping force in one direction while would bend under lateral loads. C depicts a soft gripper which is mechanically supported with the rigid bar mechanisms in order to deal with the lateral forces. Figures B and C are adopted from (Shintake et al., 2016b) and (Zhu et al., 2022) respectively .

The abovementioned soft robotics solutions collectively represent the multidimensionality in gripping options for agri-food applications. Where one solution may enhance the performance of the gripper in one dimension it tends to limit the performance in other dimensions. Therefore, we need multifaceted criteria in which all the challenges are reflected. To achieve this purpose, a performance matrix is introduced by which we can measure gripper factors, their grasping performance and variations in object handling illustrates the logic behind the criteria.

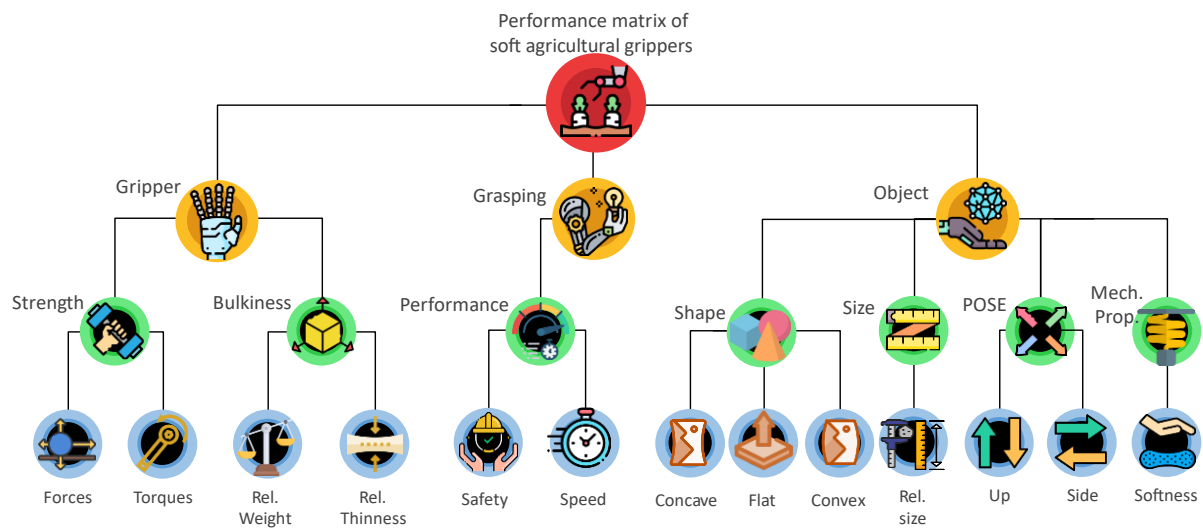

Figure 5 Shows the performance matrix for a soft robotic gripper designed for agricultural applications. It embeds three folded aspects: gripper parameters i.e. strength and bulkiness, grasping performance, and object variations. The gripper strength can be measured based on the maximum force/torques before the slippage of the object and the bulkiness could be measured by relative weight and thickness of the object to the gripper. The safety and speed are chosen as the most important indices representing the performance of grasping the crops. The shape, size, pose and the mechanical properties are the indices representing the variability of the crops/objects to be handled. (Icons are downloaded from (FlatICON, n.d.))

The gripper factors considered are strength and bulkiness which are related to the challenges 3 and 4 respectively. In particular, an ideal soft gripper for agricultural tasks should be strong enough to be able to handle loads and dynamic perturbations and be thin and agile enough to maneuver in dense vegetations (figure 5). The strength can be defined as the maximum force and torque that a gripper can handle in different directions (XYZ directions in figure 4-B). The strength factor of the gripper could be defined based on the payload to gripper's weight ratio $pwr$ in Eqn (1). The bulkiness factor could be represented by the ratio of gripper thickness $t_g$ to the object's width/thickness $t_o$ as expressed in Eqn (2).

$$pwr = \frac{Payload}{W_g} \quad (1)$$

$$tr = \frac{t_o}{t_g} \quad (2)$$

Moreover, the soft grasping should be carried out both quickly and safely so that automated harvesting is economically feasible (figure 5). Safety is a qualitative factor hence it is weighted based on an expert's opinion range from 0-1 and follows from the type of the grasping mechanisms they are using i.e. Bernoulli mechanism (Petterson et al., 2010), mechanical interlocking, (Leylavi Shoushtari et al., 2019), friction-based, controlled adhesion and controlled stiffness (Zhang et al., 2020)(Shintake et al., 2018). The Bernoulli mechanism and the adhesion-based mechanism is ranked as the safest for grippers, followed by controlled stiffness, mechanical interlocking and friction-based grasping at the other end of the spectrum of safe grasping mechanisms. The speed is also defined based on the time required for grasping.

Finally, the ideal soft gripper should be able to handle variations in crops i.e. shape, size, pose and their mechanical properties (figure 5). Shape variation can be defined as flat, convex and concave. The size can be measured based on the aspect ratio of the maximum length of the object to the length of the gripper. pose can be defined as whether the major part of the grasped object is on the side of the gripper (i.e. the weight of the object acts as a lateral load) or the grasped object is facing up or down (i.e. the gripper is either approaching from the top or from bottom). Finally, the mechanical properties of the object can be simplified as the range of the stiffness of objects that the gripper can carry. The range of the stiffness is defined based on Youngs modulus ranging from the softest to the most rigid object as grasped by the gripper. Practically, grasping a rigid object is feasible for soft grippers and the real challenge is when the object is extremely soft (softer than gripper). It means that the range of stiffnesses of the objects that the gripper can grasp depends on the Youngs modulus of the softest object grasped by the gripper. So a softness index for a gripper could be developed as the inverse of the Youngs modulus of the softest object that the gripper can handle as expressed in Eqn (3)

$$S_{index} = \frac{1}{E_{softest}} \quad (3)$$

## 2.2 Performance analysis of soft agri-food grippers

Although there are numerous soft grippers that have not been particularly developed for or integrated in agricultural takes, there are still groups of grippers or gripper principles that have the potential to be integrated in agricultural tasks. But in order to show the effectiveness of the soft grippers in agriculture, for this chapter, it is decided to only focus on those soft grippers that were designed for or tested in handling agricultural products. Therefor we analyzed a database of the soft grippers used in crop harvesting and crop handling using the above-mentioned performance matrix. This database is the same as the one used in the study of Elfferich et al., (2022). Out of 79 works that were using "soft robotic grippers" and similar keywords (check the table 1 of (Elfferich, 2022) ), 11 works which were addressing aspects mentioned in the performance matrix as described above were chosen. The grippers' scores for each parameter were then normalized with respect to the maximum scores of the given parameter for all evaluated grippers. The values of shape and pose were not normalized as they are binary parameters. The following three figures show the scores that all 11 gripper designs get for the grippers' strength and bulkiness, grasping performance, and object variation they can handle.

Figure 6-A shows the safety and speed of the selected grippers. The HASL-based gripper (Hydraulically Amplified Self-Healing electrostatic actuators) shown in figure 6-B scores as the fastest gripper (Acome et al., 2018) together with the ElectroAdhesion-based gripper shown in figure 6-C since both of these grippers are working based on the principles of electrostatic actuation which acts very fast (Shintake et al., 2016b)(Cacucciolo et al., 2019). Such an actuation enables them to do grasping in about 200 milliseconds. While the HASL-based gripper scores lower on safety (due to the friction and mechanical interlocking-based grasping), the ElectroAdhesive gripper is the safest since it uses surface adhesion to grasp objects. Figure 6-C illustrates how such a mechanism allows the gripper flaps to strongly attach to the surface of the crop, without applying any normal force to the crop which usually is the main cause of bruising and skin damage. Alongside this, a Bernoulli principle-based gripper (Petterson et al., 2010) is also identified as one of the safest grippers, since akin to the ElectroAdhesive gripper, it uses an attachment mechanism to grasp objects. This attachment mechanism works based on the pressure drop between the gripper and the object caused by positive air pressure. The advantage of this attachment mechanism is that there is always a gap between the object and gripper meaning the object is suspended resulting in *contactless grasping*.

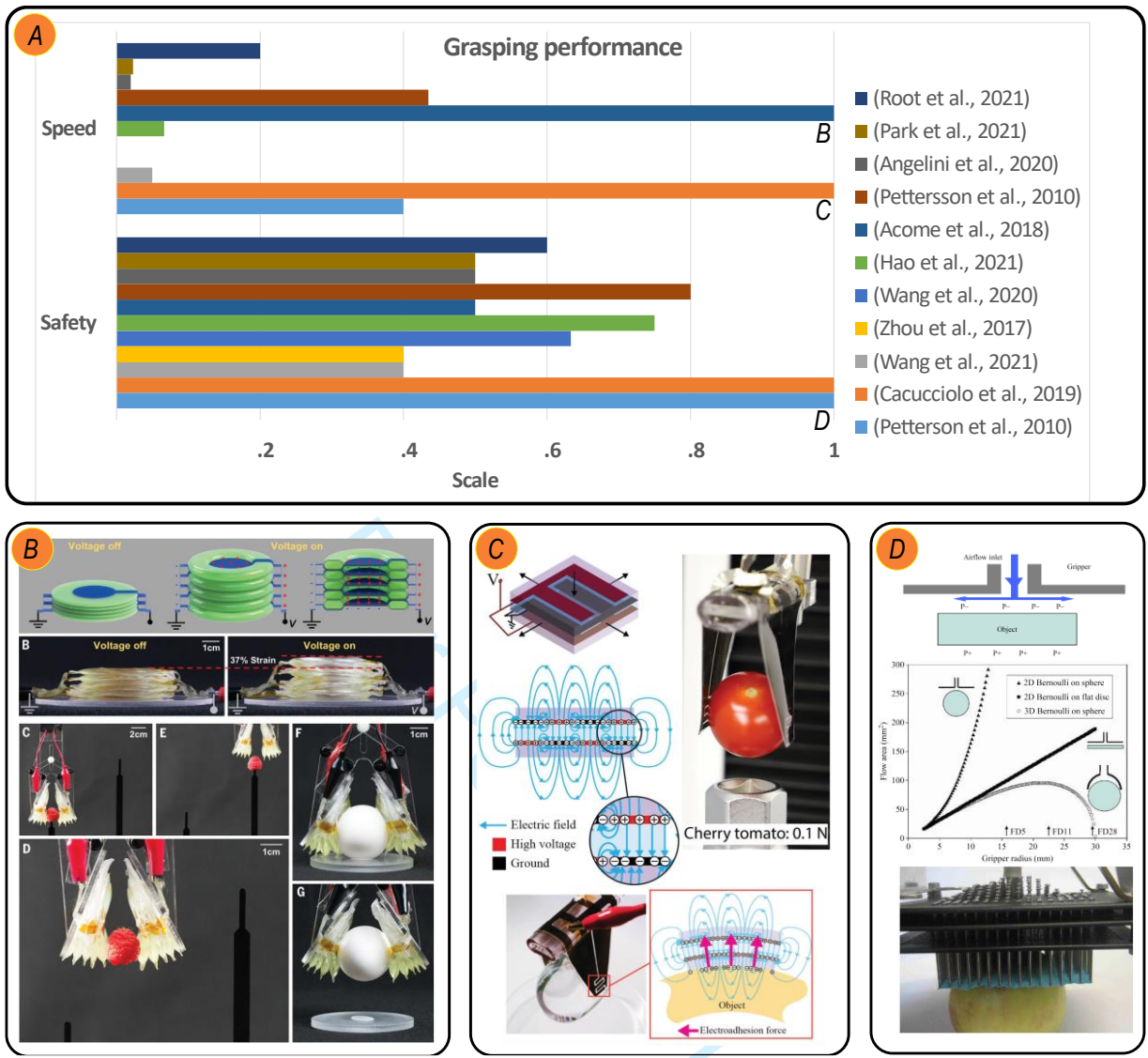

Figure 6 Grasping performance of 11 grippers. A- speed and Safety scores. B-D: grippers with the top scores. B- represents from top to bottom: HASL gripper mechanisms consisting of five ring shaped HASL actuators, the strain gained when the voltage applied, pick and placing a strawberry and an egg (adopted from (Acome et al., 2018)). C- Electro-adhesive based gripper, its mechanism and grasping time (adopted from (Cacucciolo et al., 2019) and (Shintake et al., 2016b)). D- A Bernoulli principle-based gripper, its mechanism, performance and deformation in interaction with the objects (adopted from (Petterson et al., 2010)).

The Electroadhesive gripper (Shintake et al., 2016b) is scored as *lightest* and *thinnest* gripper (Figure 7-A). It is capable of handling objects 1000 times heavier than its weight while being 172.5 times thinner than the object. Such a thin design would allow it to be integrated in picking crops from a pile where grippers need to be thin enough to get around the object. The Pisa/IIT SoftGripper is a tendon-driven anthropomorphic hand, which takes advantage of an adaptive ligament constraining system, and is scored as the strongest gripper (Figure 7-B). It is capable of handling loads in vertical direction (z) up to 200N, lateral loads up of 50N and endure torques up to 1, 0.5 and 2 Nm around x, y and z axis.

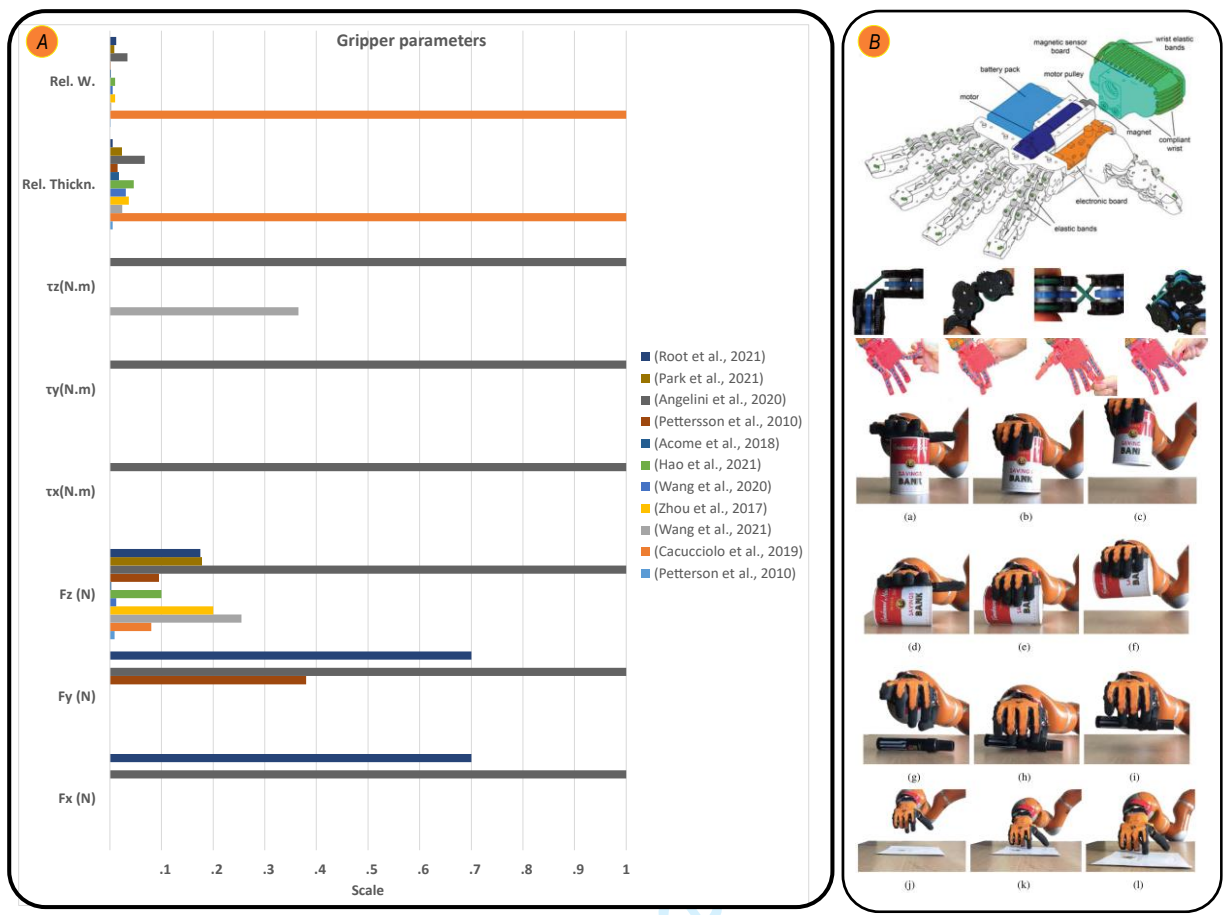

Figure 7 A Gripper parameters i.e. relative weight and thickness, and maximum forces and torques in three dimensions. B- top to bottom: the Pisa/IIT SoftGripper; the stretchable ligaments making gripper adaptive in interaction with environment; grasping different objects using a synergy control. (photo is adopted from (Catalano et al., 2014))

The figure 8 shows to what extent each soft gripper is capable of handling the object variations. The figure shows that all 11 grippers are capable of grasping objects facing up (*Up POSE)* as well as objects with *convex shapes.* Indeed, for the rest of the grippers , we can clearly see that such an experiment i.e. grasping objects that are round and facing up, are repeatedly carried out to show their performance. Hence, it is clear that these two variations i.e. Up pose and *convexity* are not challenging for soft grippers. However, soft grippers are challenged by the other object variations such as *side POSE, concave shapes,* object *softness*, *relative size* and *flat* shape.

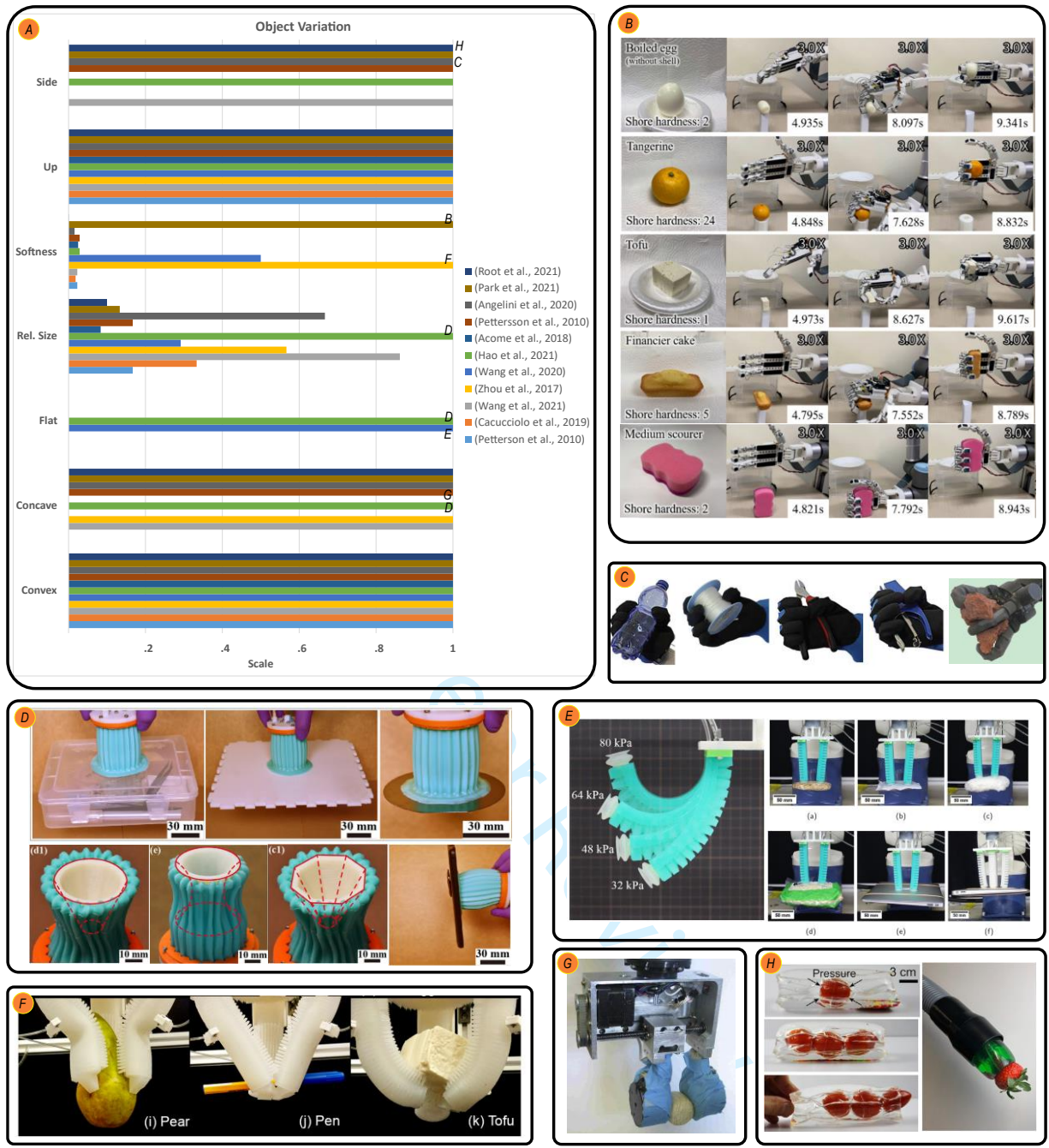

Figure 8 **A-** Shows the scores associated with the object variation. For each variation challenge factor (those for which just part of the gripper scored 1) two selected grippers with score 1 are shown with the capital letters on the bar. The capital letters refer to the visual representation of the gripper shown in B-H sections. **B-** variable stiffness anthropomorphic gripper with stiffness control, capable of grasping extra soft (Tofu), concave objects and grasp them from the side. Adopted from (Park et al., 2021). **C-**Pisa/IIT SoftGriper capable of grasping concave and unstructured objects and side grasping. Adopted from (Santina et al., 2018). **D-E** a multimodal enveloping soft gripper (Hao et al., 2021) and a dual-mode soft gripper (Wang et al., 2020) both capable of grasping concave and flat objects and grasping from the side. Images of section D and E are adopted from (Hao et al., 2021) and (Wang et al., 2020) respectively. **F-** A soft gripper with enhanced object adaptation capable of grasping concave and ultrasoft objects. The image of section F is adopted from (Zhou et al., 2017). **G-H** an adaptive magnetorheological gripper designed for food handling (Pettersson et al., 2010) and a soft bio-inspired mechanism for grasping, catching and conveying (Root et al., 2021) which both are capable of side grasping and lifting objects with concave shapes. Images of section G and H are adopted from. Images of B-H are respectively adopted from (Pettersson et al., 2010) and (Root et al., 2021) respectively.

### 2.2.1 Side POSE

All grippers with a purely soft body i.e. the Electroadhesive gripper (Cacucciolo et al., 2019), the adaptive soft hand (Zhou et al., 2017), the HASL-based gripper (Acome et al., 2018) and the dual mode soft gripper for food packaging (Wang et al., 2020) are incapable of grasping objects from the side. The reason for this is that the pure soft bodied grippers do not have enough *support stiffness*. On the other hand, the hybrid rigid-soft bodied grippers such as the soft grasping and conveying mechanism (Figure 8-H)(Root et al., 2021), the stiffness-controlled gripper (Figure 8-B) (Park et al., 2021),  the Pisa/IIT SoftGripper (Figure 8-C) (Angelini et al., 2020), the Magnetorheological gripper (Figure 8-G) (Pettersson et al., 2010), the multimodal enveloping gripper (figure 8-D) (Hao et al., 2021) and the circular shell gripper (Wang et al., 2021) are capable of side grasping. Moreover, side grasping is also a challenge for the adaptive Bernoulli gripper (Figure 6-D) (Petterson et al., 2010) for different reasons. Despite the fact that the Bernoulli grippers have been used in food industries before (Davis et al., 2008)(Erzincanli & Sharp, 1997) they have always been used in a horizontal orientation. Indeed, in this position the gripper performs best as the sheer force is minimized. Using a Bernoulli gripper for side grasping entails .

### 2.2.2 Object softness

The only two grippers that could handle objects as soft as a Tofu were using a very low stiffness profile to handle such a product (Park et al., 2021; Zhou et al., 2017). The soft segmented gripper (Zhou et al., 2017) uses an ultrasoft pillared interface to safely grasp and accommodate Tofu in a mechanical interlocking regime (Figure 8-F, on the right). Such a passive approach works until the shore hardness of the interface is lower than the object to be handled. Shore hardness is a measure of the hardness of a material, specifically its resistance to indentation or penetration by a hard object. However, if the object is relatively softer than the gripper's interface, then the grasping would be challenging.

In contrast, the anthropomorphic gripper with controlled stiffness (Park et al., 2021) can handle Tofu despite not having a soft interface by employing an adaptive stiffness approach. It enables the robotic gripper to automatically adjust the stiffness of its fingers to the stiffness and shore hardness of an unknown object once the gripper comes to interaction with it. In particular, the peak force at fingers $f_p$  is correlated with the shore hardness scale $h_s$ (of the objects) and the damping ratio $\zeta_d$ of the object as expressed by Eqns (4-5). Using the resulting correlation between peak force and shore hardness $h_s$ as expressed by Eqn (5) allows the robot to estimate the hardness of an unknown object using the force feedback received at its fingertip. Then the estimated shore hardness is used in Eqn (4) to calculate the damping ratio.

The resultant damping ratio is used by an impedance controller to adjust the stiffness of the gripper. To achieve this purpose, a list of the 21 objects with quite a different shore hardness have been used for these experiments (figure 9-A). Their standard shore hardness has been measured (figure 9-B) and used for the correlation (figure 9-C).

$$\zeta_d = \alpha_1 \, h_s^2 - \alpha_2 \, h_s + \alpha_3 \quad (4)$$

$$h_s = \beta_1 \, e^{\beta_2 f_p} \quad (5)$$

### 2.2.3 Flat objects

Once it comes to handling flat objects, only those grippers which take advantage of suction-based adhesion succeeds (Figure 8-A). Even the other type of adhesions such as Electro-adhesion or gecko-inspired dried adhesion cannot handle flat and non-lightweight objects simply because of the peeling effect (Ruotolo et al., 2021; Shintake et al., 2016a). The peeling-off effect could be solved for gecko-inspired adhesive pads by having a transmission mechanism to convert normal load (which is an effect of object weight and causes the peeling phenomenon) to shear loads which reinforces the attachments (Hawkes et al., 2016; Jiang et al., 2017).

In contrast, suction cup-based adhesion works best on flat (and non-rough) surfaces (due to the guaranteed sealing), with normal loads and might fail due to the shear forces. Indeed, the two grippers successful of grasping flat surfaces were tested in this class (figure 8-D top row and figure 8-E). Using a suction-based gripper for side grasping would be quite challenging since the weight of the object applies to the gripper as a shear load. Such a situation requires a stronger suction cup, that is equivalent to having a bigger suction cup or higher suction force, which is the case for the multimodal enveloping gripper (Hao et al., 2021) (shown in the bottom right photo of figure 8-D). It is worth mentioning that the dual mode gripper will not be able to hold flat objects as shown in figure 8-E in the side grasping scenario since it does not have a rigid support. In summary, without adhesion, it seems impossible to grasp flat or relatively big objects. This point was previously made as a design choice for versatile grippers (Langowski et al., 2020).

The Bernoulli grippers are known as contactless grippers used to handle flat objects(Brun & Melkote, 2006; Journee et al., 2011). Although they are not used widely in handling crops they still can be considered as having a great potential for the agricultural cases. One of the reasons that can explain why this integration has not taken place is that this category of grippers work optimal when the surface is perfectly flat and this usually is not the case for agricultural products. In fact, the Bernoulli-based grippers with slight modifications to deal with this issue have shown a successful results in handling non-flat objects, (Petterson et al., 2010).

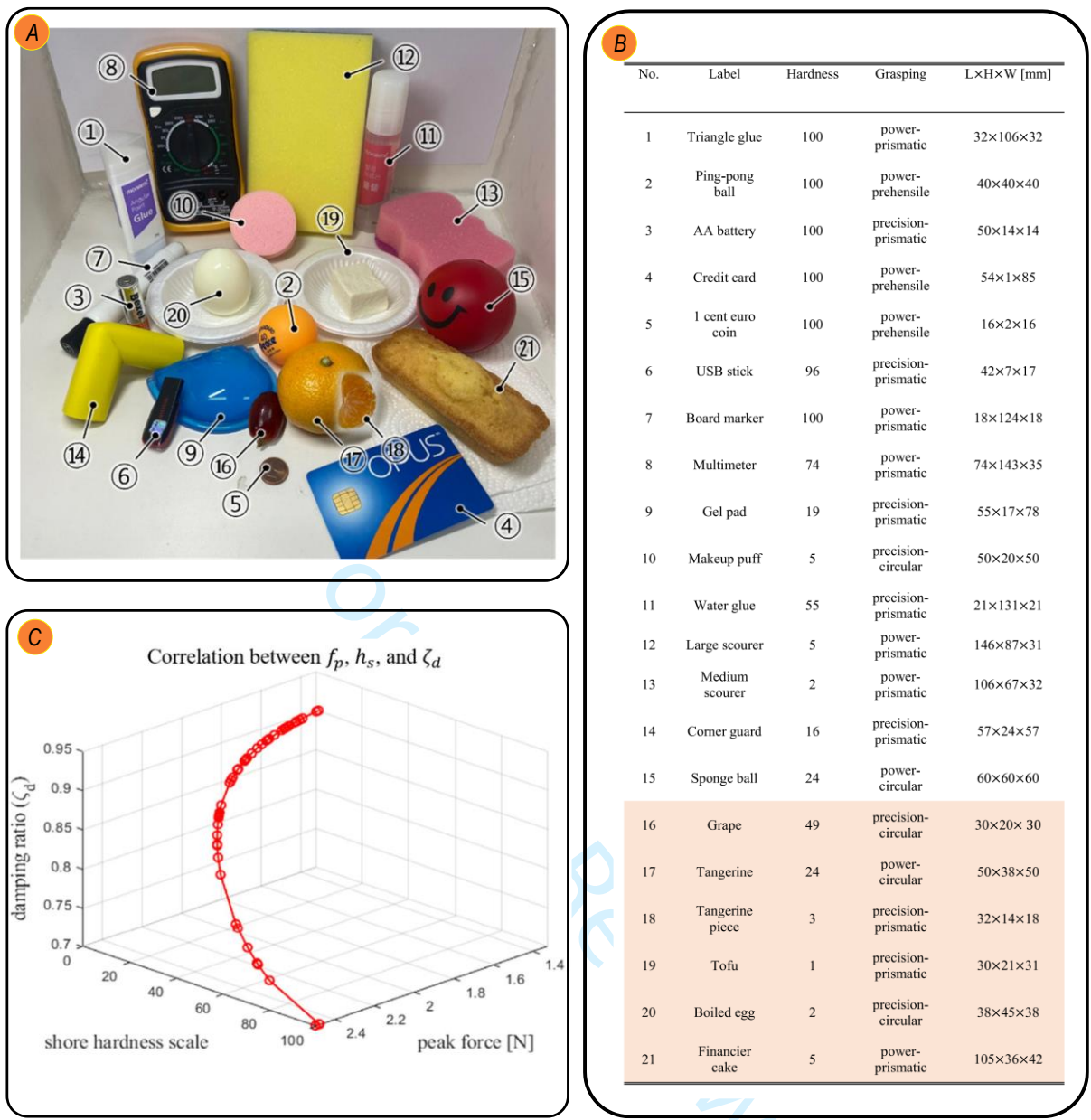

Figure 9 A- different objects used in the shore hardness-peak force and damping ratio experiment. B- their standardized shore hardness. C- correlation between the three parameters. Adopted from *(Park et al., 2021)*

### 2.2.4   Concave objects

Here we analyze the groups of grippers with successful and unsuccessful demonstration of grasping concave objects. This analysis is based on the grasping mechanism (i.e. Bernoulli principle, controlled adhesion i.e. surface adhesion or suction-base adhesion, controlled stiffness, mechanical interlocking and controlled friction) and the materials of the gripper's body (i.e. pure soft, hybrid soft-rigid) since these parameters were found to be relevant in this context. The following paragraphs are going to explain how these two factors can affect grasping of concave objects.

There is a group of four grippers without any demonstration of grasping of concave (and non-lightweight) objects i.e. the adaptive Bernoulli, ElectroAdhesive, HASL-based and the dual mode soft grippers. The adaptive Bernoulli gripper (Petterson et al., 2010) was designed to approach objects from the top and cannot reach the side of the object where the concavity is. Even though the gripper is capable of deforming and conforming to the shape of the object, the deformation is not enough to reach the side of the object. The ElectroAdhesive grasping (Cacucciolo et al., 2019) relies on a large surface contact which cannot be achieved when the object has a concave shape. In contrast, this gripper performs very well if the object is flat on the side or is convex. The other two soft grippers i.e. the dual mode gripper (Wang et al., 2020) and the HASL-based gripper, (Acome et al., 2018) both use friction-based and/or mechanical interlocking grasping mechanisms to pick up objects while having pure soft bodies.

In principle there is a group of seven grippers that demonstrated successful grasping of concave objects i.e. (Angelini et al., 2020; Hao et al., 2021; Park et al., 2021; Pettersson et al., 2010; Root et al., 2021; Wang et al., 2021; Zhou et al., 2017). Likewise the anthropomorphic gripper (Park et al., 2021) and Pisa/IIT Soft gripper (Angelini et al., 2020) are also using friction-based and mechanical interlocking grasping mechanisms while succeeding in handling concave objects. They both have a hybrid soft-rigid structure that enables them to grasp concave-shaped objects. Moreover, having the capability of stiffness control is an added value in handling such objects.

The magnetorheological gripper and the circular shelled gripper also have hybrid soft-rigid materials while using the controlled stiffness approach for grasping (Figure 8). In both cases having a rigid substrate allows their soft compartment to stiffen while they conform to the object. In particular, their rigid support allows soft pouches to be pushed to the object (during actuation) and conform to the concavity of the object which creates a large surface contact. During the stiffening phase, the grippers can apply a well-distributed force to the concave objects thanks to these large surface contacts.

The same physics underlies the grasping of toroidal hydrostat and multimodal soft grippers i.e. large surface contact followed by stiffening. The only difference is that the grippers are using pure soft bodies to conform to the object and stiffening it. The toroidal hydrostat gripper uses an inversion mechanism known as sloughing mechanism (Sadeghi et al., 2013) to drag a soft object in a ring pouch via rolling, while the multimodal gripper uses a soft shell structure to collapse in a programmed way to the object. The last gripper in this group (Zhou et al., 2017) uses a different mechanism to adapt to concave shapes. It takes advantage of a segmented finger capable of having multiple bending curvatures to conform to the objects with both concavity and convexity, such as pears (Figure 8-F). Similar to other grippers, this gripper also has a very large surface contact with objects, which helps the gripper to apply a well-distributed grasping force during the actuation.

3. **Ideal versatile gripper: future directions**

From the results so far, we can see none of the grippers are able to satisfy all the requirements associated with agricultural tasks. Each design could satisfy just one or some of the requirements and was failing on the other aspects. The reason is simple, their designs are not meant to satisfy all agricultural task requirements. Indeed as we have seen, when it comes to the design choice, soft robotics has a lot to offer despite current examples do not yet meet the aforementioned requirements. As an example, one of the challenges in grasping of objects with the side-faced pose was a lack of support stiffness which could be resolved by jamming mechanisms (Brancadoro et al., 2020; Devices & Ibrahimi, 2020) or any other soft robotic solution that offers a selective stiffness (Visentin et al., 2021). There are some examples of such a combined design aiming to compensate for the flaws of one gripper design for a certain purpose, such as increasing the strength (Yang et al., 2019; Zhou et al., 2020). However, regardless of the kind of combination of technology or gripper mechanism, *the final design will not satisfy all of the requirements* if all of the requirements are not taken into account right from the beginning.

Studying natural grippers could help us with designing the ideal versatile gripper whether via inspiration or a biomechanical principle. Indeed, organisms in their natural habitats develop functionalities in order to adapt to their environments as a generic survival approach. For example, the development of opposable thumb in chimps help them to climb thin branches to get to fruits normally growing at the tip. This could be a good candidate for a more optimal versatile gripper. Moreover, studying the neuromusculoskeletal structure of these hands would tell us much about the grasping mechanism. Researchers have shown that natural grippers use a hybrid grasping mechanism in order to deal with unpredictability and variations in their environment (Langowski et al., 2020) which is another excellent approach to enhance versatility.

This message in in line with what we have seen so far in reviewing the previously mentioned grippers: the fact that no one single grasping mechanism is able to handle the set of requirements of a natural task i.e. picking up a crop. Here we introduce a radar diagram to show how we can use all the soft robotic technologies we have reviewed so far to cover all requirements. Figure 10 illustrates such a concept where the minimum number of the complementary grasping technologies among the set of 11 grippers were chosen. Complementary grasping technologies means that when a grasping technology fails to meet some requirements, another can compensate for those requirements. These technologies are as follow:

- Electroadhesive grasping technology (showed with orange color in figure 10) which has a low bulkiness i.e. relative weight and thickness and high grasping performance i.e. safety and speed.
- The 3-finger gripper with soft palm which score high in grasping soft objects.
- The multimodal enveloping soft gripper which scored high in dealing with object variation i.e. shape size and pose.
- SoftHandler which is a tendon-based gripper scored high in strength.

Hypothetically the ideal gripper would be a hybrid design of such grasping technologies. Although, whether the hybridization of these specific technologies is feasible or not, is a design research question. This research question is particularly addressed in the section 3 case study.

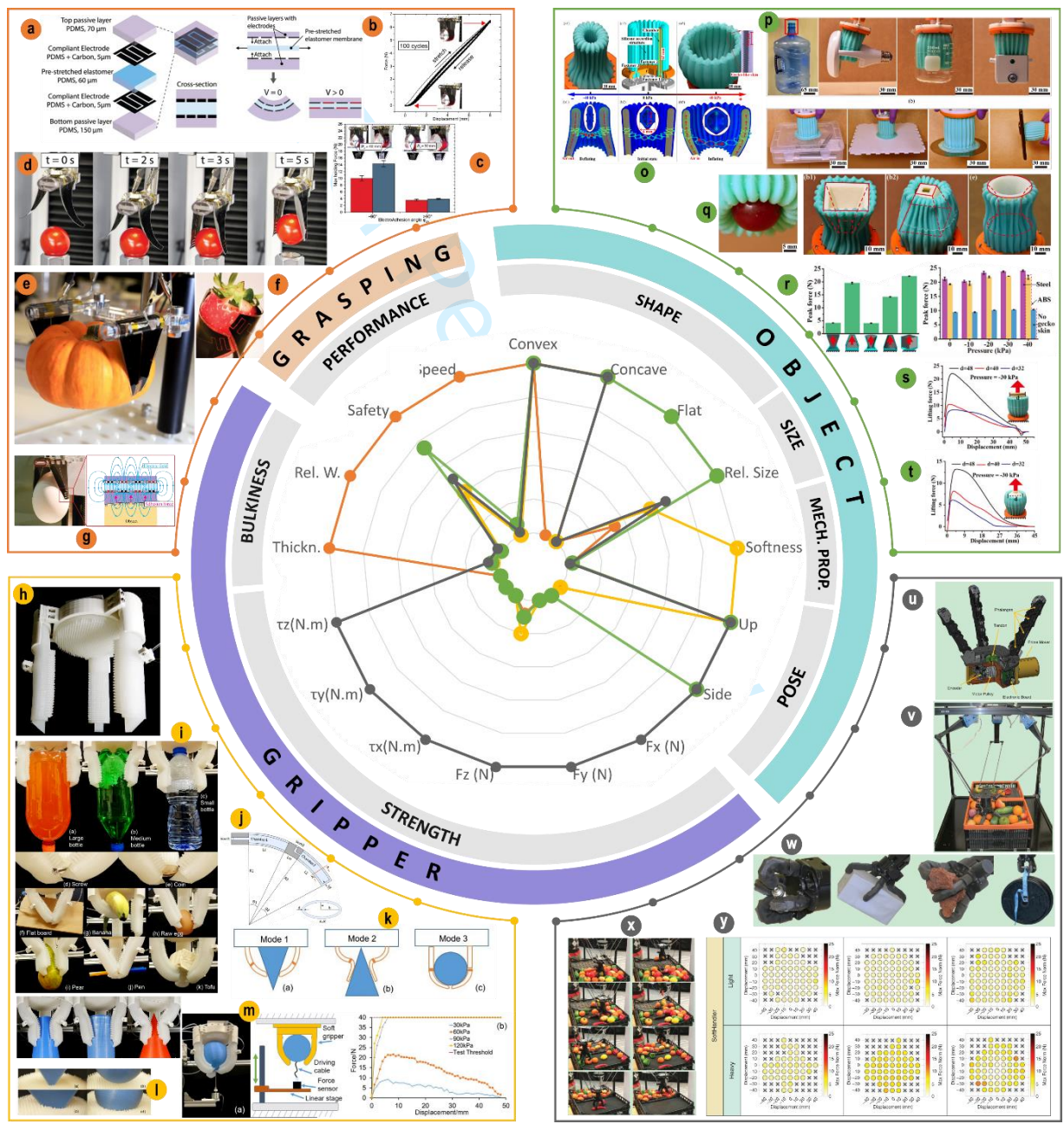

Figure 10. In the center: a graph illustrates the performance matrix measuring grippers' factors i.e. strength and bulkiness, grasping performance, and the variation of the objects that can be handled. The orange, yellow, green, and gray lines are showing four grippers with the highest scores i.e. (Cacucciolo et al., 2019), (Zhou et al., 2017), (Hao et al., 2021) and (Angelini et al., 2020) respectively. These grippers,

their grasping performance and different objects that they handled are shown in four quarters i.e. top left, bottom left, top right and bottom right respectively. The top left quarter depicts the electro-adhesion gripper: **a-** mechanism underlying the dielectric-based actuation, **b-** on-off cycle, **c-** holding force objects with different size and ElectroAdhesion angle, **d-** rapid grasping of a cherry tomato, **e-** grasping of a relatively large size pumpkin, **f-** holding a strawberry safely, and **g-** holding an egg while showing the mechanism of ElectroAdhesion. (images of sections a-d are adopted from (Cacucciolo et al., 2019)) The bottom left quarter shows **h-** a 3-finger gripper with a soft palm, **i-** holding on an object of various sizes, shapes, and weights, and ultra-soft (tofu) using the segmented finger **j** via different grasping modes (**k**). **l-m** are showing respectively the payload and the soft palm interacting with different objects. The top right quarter illustrates the multimodal enveloping soft gripper and its working principle, **o** holding a large object (with respect to the size of the gripper), a flat object and grasping object from the side **p**. **q-** shows this gripper enveloping a cherry tomato and also grasping standard objects with convex and concave shapes. **r-** performance of different grasping modalities, **s-t**: grasping performance of object with different size and shape. The bottom right illustrated the SoftHandler grasping system (**u-v**) capable of handling quite heavy and unstructured objects (**w**). **x-** pick and placing agricultural products and **y-** force distribution at gripper's palm while handling objects with various size (i.e. small, medium and large) and weights (i.e. light and heavy).

## 4. Case study

A very good use-case of the applications of soft robotics in agriculture is the harvesting in intercropping system as it is a case with a wide-set of requirements. The case is taken from the project 'Universal Soft Robotic Harvester for autonomous intercropping system (USOROH) (Leylavi Shoushtari & van Henten, 2020) and is a collaboration between Wageningen University and China Agricultural University as part of the Agricultural Green Development Plan. The purpose of this project is to enhance the versatility of the soft grippers (through hybrid design) to deal with the high variation of the crops in intercropping system. In particular, versatility is defined based on the adaptability to the shape, size, weight and mechanical properties of the objects to be handled. Each soft grasping technology has a list of pros and cons regarding the adaptability, which through a hybrid design their drawbacks can be traded-off or ideally canceled-out.

Intercropping is a farming practice in which a set of crops with mutual benefits are planted next to each other which results in using less to no pesticides, fertilizers and enhances the soil health (Vandermeer, 1989). Having different crops in the same field is always a challenge when it comes to harvesting since it has both categories of diversities: the diversity within a crop variant and the diversity between different types of crops. Agricultural technologies are mainly designed for monocultures that are shown to be unsustainable. Whereas traditional farming practices such as intercropping or mixed cropping have been proven to be more sustainable, particularly for soil health. Such farming practices are abandoned mainly due to the intensive labor requirements for planning, crop maintenance and harvesting.

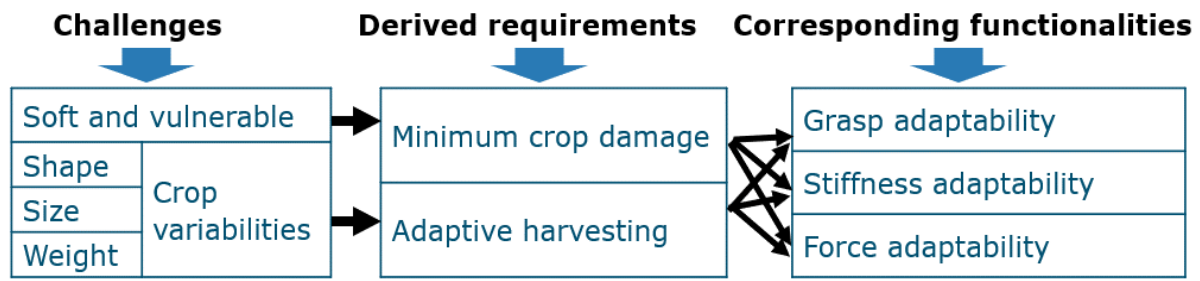

Figure 11 The chain of challenges, requirements and functionalities for harvesting task in an intercropping system.

Following a systematic approach (Figure 11), a set of functional requirements of a gripper (i.e. grasping, stiffness and force adaptability) can be driven from the challenges of harvesting in an intercropping farm. Current soft grasping technologies could provide these functionalities individually but not simultaneously. Therefore, the existed soft grasping technologies are assessed based on the abovementioned functionalities and a set of candidate technologies are chosen for hybridization. As an example, three different grasping technologies i.e. controlled stiffness, controlled adhesion, and grasping by actuation have been addressed (Shintake et al., 2018). Our aim is to combine these different technologies to conceptualize a design for the ideal versatile gripper. Knowing that each of these technologies has its own pros and cons, we try to group the candidate technologies based on the task requirements mentioned in this chapter.

5. Summary

There is a high demand for the automation of harvesting since it takes a remarkable portion of the production cost, about one third of the entire cost for producing valuable crops (Bac et al., 2014). Despite this fact, current rigid robots and machines and automation technologies are not ready to meet the high requirements of mixed-crop harvesting.

This chapter aimed to address that the integration of the current soft grasping technologies in agri-food tasks (particularly the case of harvesting) is a multifaceted research problem. It requires to simultaneously take the following aspects into account: gripper parameters (i.e. strength and bulkiness) grasping performance, and object variations. Therefore, to see to what extend the soft grasping technologies are capable to be integrated in harvesting case, these triple aspects of each technology required to be assessed. This chapter provide with an assessment matrix embedding the abovementioned three-folded aspects. The soft grippers are assessed based on two parameters i.e., strength and bulkiness. The former could be measured based on the maximum force/torques before the slippage of the object and the latter could be measured by relative weight and thickness of the object to the gripper. The grasping performance is assessed based on the safety and speed as the most relevant factors in crop handling. The shape, size, pose and the mechanical properties are the indices representing the variability of the crops/objects to be handled. Subsequently this chapter provide with numeric indices of the assessment matrix. There were two main challenges in extracting data from literature:

- There are not enough data about the assessment matrix.
- Some measures are qualitative i.e., safety, shape and roughness which makes it difficult for benchmarking.

The results of the comparison of different grasping technologies showed the grippers with respect to each other. These results also reveal the fact that certain features of the objects (i.e. flat and concave shapes, softness and side pose) are quite challenging for grippers as just few grippers were scoring high. It also has been discussed why these grippers are outperforming their counterparts in such challenging cases.

In section 3 a hypothetical design of an ideal gripper is presented for the harvesting task. In fact, it is a hybrid design of soft grasping technologies that complement each other regarding the assessment matrix represented in section 2.1. In section 4 a case of Universal Soft Robotic Harvester for autonomous intercropping system (USOROH) was represented aim to reflect the application of such a hybrid design. In broder context, soft robotics could provide a more safe, flexible and hence inclusive solution for harvesting of the mixed crop case. Although, the safety that has been discussed here only is about maintaining crop quality during robotic handling, but soft robots also provide a quite safe interaction with humans too. I-support is a good (non-agricultural) example for demonstrating soft robots' safety in human-robot interaction (Arleo et al., 2020; Zlatintsi et al., 2020). Moreover, soft robotics unlocks access to a library of affordable and available materials impossible with rigid robots (Rus & Tolley, 2015). Moreover, having access to certain manufacturing technologies i.e. 3D printing and casting makes it even easier to design, customize or reproduce a soft robot. Platforms such as the Soft Robotics toolkit (Soft Robotics Toolkit, n.d.) and FlowIO (Shtarbanov, 2021) are very good examples of open access soft robotics helping to democratize the soft robotics as a solution for all users and specially farmers.

Therefore, soft robotics could potentially offer not only safe and versatile but also affordable and accessible solutions for harvesting of mixed crop farms. This chapter aimed to address the technological gap between the actual performance of soft robotics and its potential for agriculture and particularly for sustainable farming practices such as intercropping. In order to turn this potential to action, we need further research such as the USOROH project to take the maximum advantage of current soft robotic technologies and sciences to develop tailored solutions for sustainable farming practices.

## 6. Where to look for further information

The following articles provide a good overview of the subject:

Elfferich, J. F., Dodou, D., & Della Santina, C. (2022). Soft Robotic Grippers for Crop Handling or Harvesting : A Review. *IEEE Access*, *10*(June), 75428–75443. https://doi.org/10.1109/ACCESS.2022.3190863

Shintake, J., Cacucciolo, V., Floreano, D., & Shea, H. (2018). Soft Robotic Grippers. *Advanced Materials*, *30*(29). https://doi.org/10.1002/adma.201707035

Langowski, J. K. A., Sharma, P., & Leylavi Shoushtari, A. (2020). In the soft grip of nature. *Science Robotics*, *5*(49), 3–6. https://doi.org/10.1126/scirobotics.abd9120

Bac, C.W., van Henten, E.J., Hemming, J. and Edan, Y. (2014), Harvesting Robots for High-value Crops: State-of-the-art Review and Challenges Ahead. J. Field Robotics, 31: 888-911. https://doi.org/10.1002/rob.21525 Koerhuis, R. (n.d.). Wageningen University on a quest to develop the best robotic grippers. *Future Farming*. https://www.futurefarming.com/crop-solutions/wageningen-university-on-a-quest-to-develop-the-best-robotic-grippers/

Zhang, B., Xie, Y., Zhou, J., Wang, K., & Zhang, Z. (2020). State-of-the-art robotic grippers, grasping and control strategies, as well as their applications in agricultural robots: A review. *Computers and Electronics in Agriculture*, *177*(April), 105694. https://doi.org/10.1016/j.compag.2020.105694

Magazine of Future Farming:

https://www.futurefarming.com/crop-solutions/wageningen-university-on-a-quest-to-develop-the-best-robotic-grippers/

Website of Dutch Soft Robotics Community:

https://dutchsoftrobotics.nl/

Open source soft robotics technologies:

https://softroboticstoolkit.com/

https://www.softrobotics.io/

http://opensoftmachines.com/